\documentclass[10pt,twocolumn,letterpaper]{article}

\usepackage{wacv}              

\usepackage{multirow}
\usepackage{placeins} 

\definecolor{wacvblue}{rgb}{0.21,0.49,0.74}
\usepackage[pagebackref,breaklinks,colorlinks,allcolors=wacvblue]{hyperref}

\def\wacvPaperID{3298} 
\def\confName{WACV}
\def\confYear{2026}

\title{ContextAnyone: Context-Aware Diffusion for Character-Consistent Text-to-Video Generation}

\author{Ziyang Mai\\
Dartmouth College\\
\and
Yu-Wing Tai\\
Dartmouth College\\
}

\begin{document}
\maketitle

\begin{abstract}
Text-to-video generation has advanced rapidly, yet preserving a character's holistic appearance from a single reference image remains challenging, particularly when the character undergoes large pose, motion, and scene changes. Existing reference-conditioned approaches primarily treat the reference image as a conditioning signal, which can weaken fine-grained appearance information as reference and noisy video tokens interact during denoising. We propose \textbf{ContextAnyone}, a context-aware diffusion framework that instead treats the reference as an explicitly preserved appearance anchor. Our key idea is to jointly reconstruct the reference image and generate the target video within a shared diffusion transformer, providing direct supervision for preserving identity and fine-grained appearance throughout denoising. To maintain the reference as a stable source of appearance information, we further introduce asymmetric information flow that allows video tokens to selectively access reference tokens while preventing noisy video features from propagating back to the reference branch. We complement this design with Gap-RoPE, which separates the positional representations of the reference and generated video tokens. Experiments on a benchmark constructed from OpenVid-HD demonstrate that ContextAnyone improves both identity and fine-grained appearance consistency over existing reference-conditioned baselines while maintaining motion characteristics close to the underlying text-to-video generator. 
\end{abstract}

\begin{figure*}[t]
  \centering
  \includegraphics[width=\textwidth]{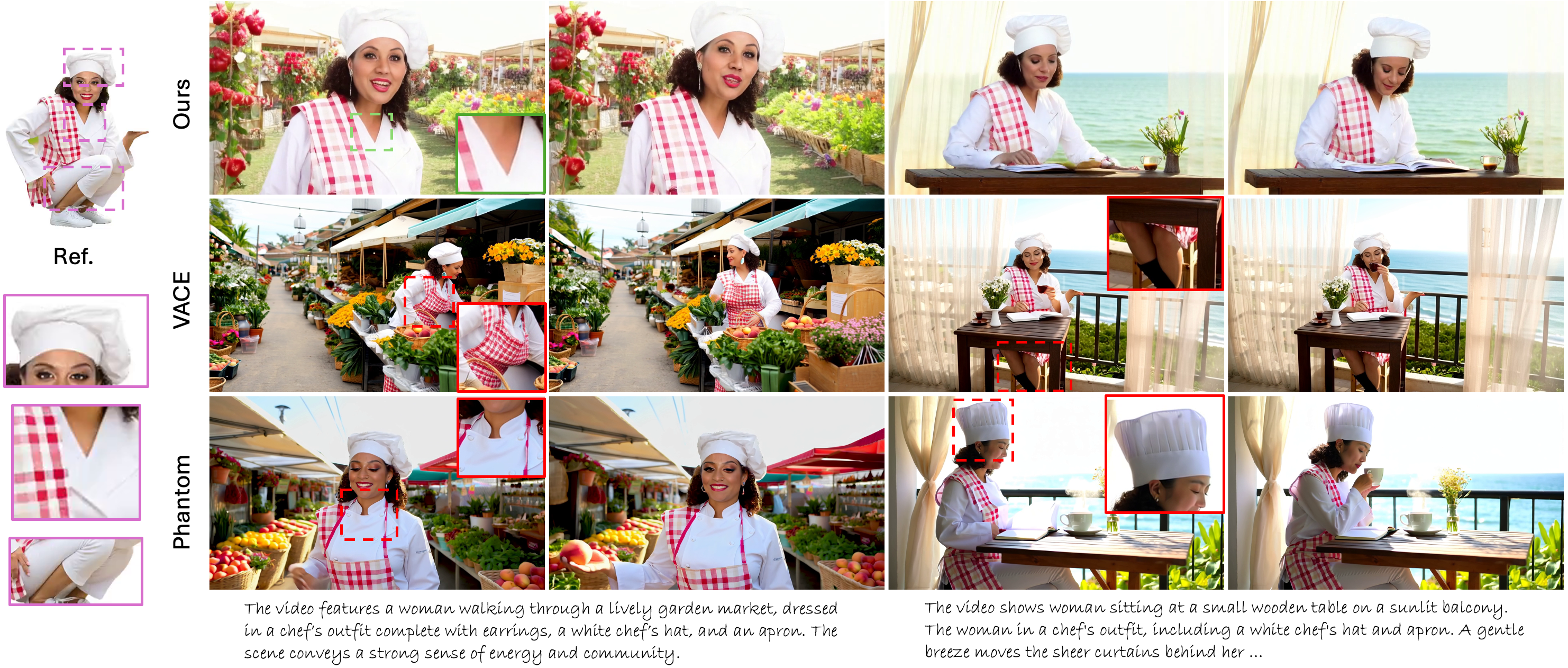}\\
  \vspace{-0.1in}
  \caption{Overview of \textbf{ContextAnyone}. Given a reference image and a text prompt, our model generates character-consistent videos that preserve visual details across diverse scenes, while prior methods struggle to retain all elements from the reference. Pink boxes highlight key details such as the chef hat, collar shape, and pant. Green boxes indicate regions where these details are faithfully preserved, whereas red boxes mark inconsistencies, such as the collar mismatch in the lower left. Many other inconsistencies are omitted for simplicity.}
  \vspace{-0.1in}
  \label{fig:teaser}
\end{figure*}

\section{Introduction}
\label{sec:intro}

Recent advances in text-to-video (T2V) generation\cite{chen2023videocrafter1,hong2022cogvideo,yang2024cogvideox,chen2024videocrafter2,genmo2024mochi,kong2024hunyuanvideo,wan2025wan,wang2023modelscope} 
have enabled realistic and temporally coherent video synthesis. However, maintaining \textit{character consistency} remains challenging, particularly for person-centric generation, where identity extends beyond facial features to hairstyle, clothing, and body shape. As illustrated in the second and third rows of Figure~\ref{fig:teaser}, existing methods often fail to preserve these contextual details across different poses and scenes, resulting in visible appearance inconsistency and reduced realism.

Existing personalization methods only partially address this challenge.
Many prior approaches~\cite{yuan2025identity,zhang2025magicmirror,li2025personalvideo,zhong2025concat} focus primarily on facial identity transfer by injecting features extracted from face encoders, which helps preserve identity at the face level but often neglects contextual appearance cues beyond the face.
More recent reference-conditioned methods~\cite{jiang2025vace,liu2025phantom,deng2025magref,fei2025skyreels} incorporate richer appearance information through pixel-level or channel-level fusion. Nevertheless, when reference and video representations are jointly processed during diffusion, the reference is typically treated as a conditioning input rather than as an explicitly preserved representation. As a result, the appearance information needed to maintain a character's identity and contextual details may become less stable during denoising, leading to identity drift and inconsistent appearance.

This observation motivates a simple question: \textbf{can the reference itself be explicitly preserved while it is used to guide video generation?} We argue that the reference should be treated not merely as a source of conditioning information, but as an \textit{appearance anchor} that remains stable throughout the denoising process. Based on this perspective, we propose \textbf{ContextAnyone}, a context-aware diffusion framework for character-consistent text-to-video generation. Instead of using the reference only to condition the generated video, ContextAnyone jointly processes the reference and noisy video latents within a shared diffusion transformer and explicitly reconstructs the reference. This additional reconstruction objective directly supervises the model to retain the reference's appearance information while learning to generate the target video.

Joint reconstruction alone, however, does not guarantee a stable reference representation. In standard self-attention, reference and noisy video tokens can exchange information bidirectionally, allowing noisy video features to weaken the appearance anchor used for conditioning. We therefore introduce an asymmetric information-flow design that allows video tokens to access reference tokens while suppressing the reverse interaction. In addition, because the reference image and generated video have different semantic roles rather than representing consecutive video frames, we introduce Gap-RoPE, which applies a fixed positional offset between the two token streams to explicitly separate their temporal positions.

Together, these designs establish a reference-preserving generation process: the reference is explicitly supervised as an appearance target, protected from contamination by noisy video features, and distinguished positionally from the generated temporal sequence. Importantly, these components are complementary rather than independent objectives: joint reconstruction provides the preservation signal, asymmetric attention protects the reference information flow, and Gap-RoPE reduces the positional ambiguity between the reference and generated video.

On a Wan 2.1 1.3B backbone, ContextAnyone achieves the strongest overall reference-related consistency among the evaluated public baselines, improving facial identity and fine-grained appearance while retaining motion statistics close to the underlying generator (Figure~\ref{fig:teaser}). A human preference study supports the improved appearance consistency, while ablations show that each component contributes to different aspects of video quality.

\noindent\textbf{Our main contributions are summarized as follows:}
\begin{itemize}[leftmargin=1.2em,itemsep=2pt,topsep=2pt]
\item \textbf{Reference-preserving generation.} We formulate reference-conditioned video generation as a joint reconstruction and generation problem, treating the reference image as an explicitly supervised appearance anchor rather than only as a conditioning signal.

\item \textbf{Protected reference-to-video information flow.} We introduce an asymmetric attention design that strengthens the transfer of reference information to video tokens while preventing noisy video features from modifying the reference representation.

\item \textbf{Positional separation for reference-conditioned diffusion.} We propose Gap-RoPE, a lightweight positional design that explicitly separates the reference and generated video token streams.

\item \textbf{Comprehensive evaluation.} We construct a challenging single-reference benchmark and demonstrate improved identity and fine-grained appearance consistency over representative public baselines while maintaining the motion characteristics of the underlying video generator.
\end{itemize}

\section{Related Works}
\label{sec:relatedwork}

\noindent\textbf{Diffusion Models for Video Generation} Diffusion models have recently driven major advances in text-to-video (T2V) synthesis.
Early methods~\cite{khachatryan2023text2video, guo2023animatediff, singer2022make, an2023latent} extend pretrained image diffusion frameworks such as Stable Diffusion~\cite{rombach2022high} by inserting lightweight temporal attention or motion adapter layers~\cite{houlsby2019parameter}.
These extensions reuse strong spatial priors learned from large image datasets, enabling short video generation with minimal retraining.
In contrast, training-based video diffusion models~\cite{wang2023modelscope, chen2023videocrafter1, chen2024videocrafter2, blattmann2023align} employ spatio-temporal U-Nets with 3D convolutions or dedicated temporal attention modules to directly capture motion dynamics from video data.
While effective, such models are computationally expensive and difficult to scale.

Recent research has shifted toward transformer-based diffusion architectures.
Diffusion Transformers (DiT)~\cite{peebles2023scalable} have become the dominant design for high-quality video generation~\cite{ma2025latte, zheng2024open, hacohen2024ltx, yang2024cogvideox, hong2022cogvideo}.
Large-scale DiT models such as Wan~\cite{wan2025wan}, HunyuanVideo~\cite{kong2024hunyuanvideo}, and Mochi~\cite{genmo2024mochi} demonstrate excellent scalability and visual fidelity compared with U-Net methods.
Our work follows this trend by adopting a DiT-based backbone for its strong generation capability and efficiency, but extends it with context-aware conditioning and attention modulation that explicitly preserve subject identity and contextual consistency across frames, which standard DiT models lack.

\vspace{2mm}
\noindent\textbf{Reference-Conditioned Video Generation} Reference-to-video (R2V) generation focuses on synthesizing personalized videos based on a given reference image or video.
Earlier methods~\cite{huang2025videomage, wu2025customcrafter, wang2024customvideo, wei2024dreamvideo} adapt pretrained diffusion models to each subject through test-time fine-tuning.
For instance, VideoMage~\cite{huang2025videomage} trains LoRA~\cite{hu2022lora} modules for each reference input before inference.
Although such methods can produce identity alignment, they are computationally demanding and unsuitable for real-time or large-scale scenarios due to the need for per-instance optimization.

To overcome these constraints, more recent approaches~\cite{jiang2025vace, liu2025phantom, chen2025multi, cai2025omnivcus, huang2025conceptmaster, yuan2025identity, deng2025cinema, fei2025skyreels} integrate reference conditioning directly into the model architecture.
For instance, ConsisID~\cite{yuan2025identity} employs frequency-domain decomposition to preserve facial identity, yet its focus remains limited to facial identity.
SkyReels-A2~\cite{fei2025skyreels} extends video diffusion transformers with channel-wise composition for element-to-video generation, but relies on a simple masking strategy that copies mask frames across time, weakening backbone modeling capacity~\cite{deng2025magref} and causing instability and identity drift.
Our method addresses these limitations with a unified, context-aware framework that maintains both global identity and fine-grained appearance from a single reference image, strengthening reference features during denoising and stabilizing temporal attention to prevent identity drift.

\section{Method}

\subsection{Architectural Challenges and Design Rationale}
\label{sec:design_principles}

We focus on DiT-based R2V models that concatenate reference and video latents into a shared sequence, as adopted by recent methods such as VACE~\cite{jiang2025vace} and Phantom~\cite{liu2025phantom}. This design provides a direct path for transferring dense appearance information from the reference to the generated video. In contrast, methods such as ConsisID~\cite{yuan2025identity} inject features from a separate face encoder and primarily emphasize facial identity, potentially capturing less information about clothing, accessories, and other appearance cues. Our analysis therefore focuses on the token-concatenation family.

In this setting, we identify three challenges that arise when the reference is jointly processed with noisy video tokens: reference preservation, asymmetric information flow, and positional separation. These challenges motivate the three components of ContextAnyone: joint reconstruction, asymmetric attention, and Gap-RoPE, respectively.

\noindent\textbf{Reference Preservation.} When the reference is used only as a conditioning input, its appearance is supervised only indirectly through the video-generation objective. Fine-grained details such as clothing, accessories, and hairstyle may therefore not be explicitly preserved. We introduce joint reference reconstruction to directly supervise the reference branch and encourage it to retain the appearance information needed for consistent generation.

\noindent\textbf{Asymmetric Information Flow.} Standard self-attention allows bidirectional interactions between reference and video tokens. Since video tokens are noisy and continuously evolve during denoising, allowing them to modify the reference representation may weaken the appearance anchor used for conditioning. We therefore enforce asymmetric information flow, allowing video tokens to access reference tokens while suppressing the reverse interaction.

\noindent\textbf{Positional Separation.} The reference image and generated video have different semantic roles, yet a standard continuous positional encoding treats them as adjacent elements of the same temporal sequence. We therefore introduce Gap-RoPE to explicitly separate their positional representations and reduce this ambiguity.

These designs work together to preserve the reference as a stable appearance anchor while enabling its information to guide video generation.

\begin{figure*}[t]
    \centering
    \includegraphics[width=1.0\linewidth]{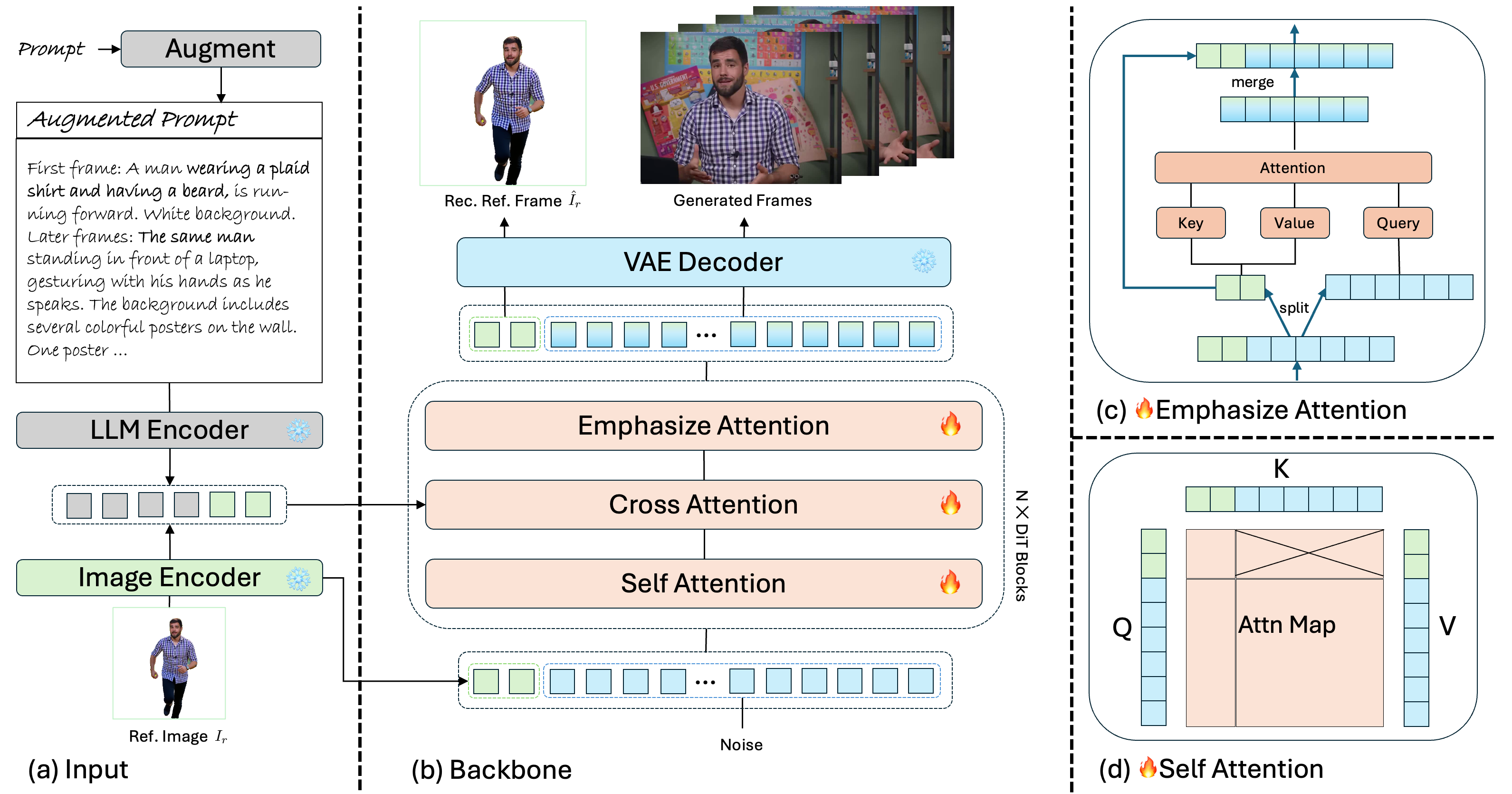}
    \caption{Framework of ContextAnyone.
\textbf{(a) Input:} The model takes a text prompt and a reference image $I_r$. The prompt is augmented by a VLM and encoded by an LLM encoder, while the image is processed by two encoders: one for cross-attention guidance and one for latent concatenation.
\textbf{(b) Backbone:} The DiT backbone contains stacked blocks with three attention modules. The input latent merges reference latents (green) and noisy video latents (blue), which are decoded separately by the VAE after the DiT blocks.
\textbf{(c) Emphasize Attention:} Latents are split into reference and video parts; video latents serve as queries and reference latents as keys and values to reinforce identity.
\textbf{(d) Self-Attention:} The attention map is masked so that reference latents do not query video latents, enforcing a one-way information flow from reference to video tokens.
}\vspace{-0.15in}
    \label{fig:method}
\end{figure*}

\subsection{ContextAnyone Framework}
\label{sec:framework}

\textbf{Problem Formulation.} Given a textual description $T$ and a reference image $I_r$ of the target person, our goal is to generate a temporally coherent video $V = \{f_t\}_{t=1}^{T_f}$ that preserves both identity and contextual appearance (e.g., hairstyle, outfit, and body shape). Figure~\ref{fig:method} illustrates the joint-reconstruction and asymmetric-attention designs; Sec.~\ref{sec:gap_rope} describes Gap-RoPE.

\noindent \textbf{Reference Encoding.} As illustrated in Fig.~\ref{fig:method}(a), we encode the conditioning signals through complementary pathways. The reference image $I_r$ is processed by two image pathways: a CLIP image encoder extracts high-level semantic identity features for cross-attention guidance, while a VAE encoder maps $I_r$ into a dense latent $z_r$ that preserves fine-grained visual details. The former provides global semantic guidance, and the latter supplies spatially aligned appearance evidence for joint denoising.

\noindent \textbf{Joint Reconstruction for Reference Preservation.}
We treat the reference image not only as a conditioning signal, but also as an explicit reconstruction target, jointly reconstructing the reference frame while denoising the video latents in a shared DiT backbone.
Specifically, we concatenate the reference latent $z_r$ with the noised video latent $z_t^v$ and feed the combined sequence into the backbone. As shown in Fig.~\ref{fig:method}(b), the reference latents (green) and noisy video latents (blue) are jointly processed by DiT blocks. After denoising, the output tokens are split into the reconstructed reference latent $\hat{z}_r$ and the denoised video latent $\hat{z}^v$, and both branches are decoded separately by the VAE.
Noise is applied only to the video branch, while $z_r$ remains clean. The reconstruction objective therefore asks the shared DiT blocks to preserve a stable reference representation in the presence of noisy video tokens. During inference, we restore the clean $z_r$ at every denoising step, preventing errors accumulated in the reference branch from being fed back into subsequent steps.
We supervise the reference branch with a reconstruction loss:
\begin{equation}
\mathcal{L}_{\text{ref}} = \| \hat{I}_r - I_r \|_2^2,
\end{equation}
and the video branch with the standard diffusion objective:

\begin{equation}
\mathcal{L}_{\text{gen}} = \mathbb{E}_{\epsilon, z_t, t}\left[\| \epsilon - \epsilon_\theta(z_t, t) \|_2^2 \right],
\end{equation}
The total training loss is defined as
\begin{equation}
\mathcal{L}_{\text{total}} = \mathcal{L}_{\text{gen}} + \lambda \, \mathcal{L}_{\text{ref}},
\end{equation}
where $\lambda$ controls the strength of the identity reconstruction constraint.

\noindent \textbf{Asymmetric Attention Design.}
We use two complementary attention modifications inside each DiT block, illustrated in Fig.~\ref{fig:method}(c)(d): Emphasize Attention strengthens reference-to-video transfer, and a directional self-attention mask blocks the reverse path.

\textit{Emphasize Attention.} Standard cross-attention in the DiT blocks is insufficient to preserve reference-specific appearance during denoising.
We therefore introduce an additional \textit{Emphasize Attention} module inserted after cross-attention to explicitly re-inject reference information into video tokens. Let $H_r$ and $H_v$ denote the reference and video tokens, respectively. We update only the video tokens via
\begin{equation}
H_v' = H_v + \mathrm{Attn}(W_q H_v,\, W_k H_r,\, W_v H_r),
\end{equation}

where video tokens serve as queries and reference tokens serve as keys and values. As shown in Fig.~\ref{fig:method}(c), the updated video tokens are then merged back with the unchanged reference tokens for subsequent diffusion blocks. This asymmetric interaction reinforces appearance transfer from the reference branch to the video branch.
This targeted interaction selectively integrates reference-aware features into the video representation, improving identity alignment and temporal consistency.

\textit{Directional Self-Attention Mask.}
In conventional self-attention, all tokens attend to each other bidirectionally.
However, this unrestricted interaction can corrupt the reconstructed reference representation, since reference tokens may absorb irrelevant motion cues from noisy video tokens.
To avoid this, we impose a directional mask on self-attention. As shown in Fig.~\ref{fig:method}(d), we mask the entries where reference tokens act as queries and video tokens act as keys, while still allowing video tokens to attend to reference tokens. This asymmetric constraint preserves the reference branch as a stable appearance source and enables one-way information flow from reference tokens to video tokens.
Writing the concatenated sequence as $H=[H_r;H_v]$, let $\mathcal{R}$ and $\mathcal{V}$ denote the reference and video token index sets. We define the mask and the resulting self-attention as
\begin{equation}
\begin{aligned}
M_{ij}&=\begin{cases}
-\infty, & i\in\mathcal{R},\ j\in\mathcal{V},\\
0, & \text{otherwise},
\end{cases}\\
\mathrm{SA}(H)&=\mathrm{Softmax}\!\left(\frac{QK^\top}{\sqrt{d}}+M\right)V.
\end{aligned}
\end{equation}
where $i$ indexes query tokens and $j$ indexes key tokens. Thus, only the reference-query and video-key block is suppressed; video queries can still aggregate both reference and video evidence.

\noindent \textbf{Text Augmentation (Complementary Input-Side Design).}
We treat prompt augmentation as an input-side complement rather than one of the three architectural contributions. As shown in Fig.~\ref{fig:method}(a), we split the input into a first-frame prompt and a later-frame prompt before encoding it with the LLM. The former describes the subject's appearance and is prefixed with \textit{First frame:}; the latter begins with \textit{Later frame: The same person ...} and describes motion and scene content. This decomposition provides structured textual guidance aligned with joint reference reconstruction and subsequent video generation.

\begin{figure}
    \centering
    \includegraphics[width=1.0\linewidth]{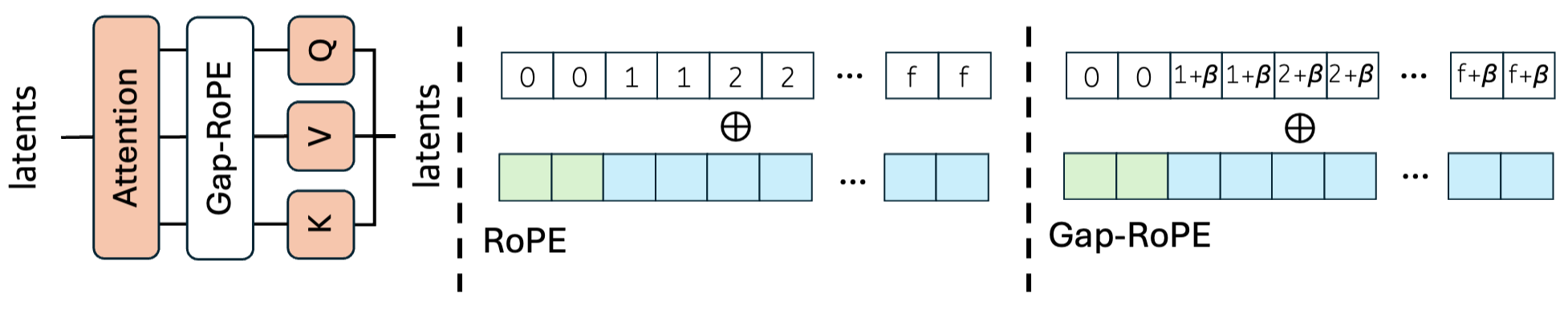}\\
    \vspace{-0.1in}
    \caption{
    \textbf{Left:} Architecture of self-attention with Gap-RoPE applied to Q and K.
    \textbf{Mid:} Standard RoPE assigns continuous temporal indices across all tokens.
    \textbf{Right:} Gap-RoPE introduces a shift $\beta$ from the generated video frame onward, creating a positional gap between reference and video tokens.
}\vspace{-0.1in}

    \label{fig:gaprope}
\end{figure}

\subsection{Gap-RoPE: Positional Separation}
\label{sec:gap_rope}

Rotary Position Embedding (\textbf{RoPE})~\cite{su2024roformer} is commonly used in self-attention to encode spatial and temporal order among latent tokens.
Standard RoPE assumes that all frame tokens belong to a single temporally continuous sequence.
In our framework, however, the token sequence is composed of two semantically different segments: a reconstructed reference segment and a generated video segment.
Treating them as a single continuous timeline introduces positional ambiguity at the boundary between reference and video, which can interfere with temporal modeling during joint reconstruction and generation.

To alleviate this mismatch, we introduce \textit{Gap-RoPE}, which inserts an explicit positional gap between reference tokens and generated video tokens. As illustrated in Figure~\ref{fig:gaprope}, standard RoPE assigns continuous temporal indices across all tokens (upper right), whereas Gap-RoPE shifts the positional indices of the generated segment by a fixed offset $\beta$ (lower right).
Formally, let $t_i$ be the standard temporal index of token $i$ and let $R(p)$ denote the RoPE rotation at position $p$. We define
\begin{equation}
p_i=t_i+\beta g_i,\qquad
\widetilde{q}_i=R(p_i)q_i,\qquad
\widetilde{k}_i=R(p_i)k_i,
\end{equation}
where $g_i = 0$ for reference tokens and $g_i = 1$ for generated tokens. Attention is computed with $\widetilde{q}_i$ and $\widetilde{k}_i$, while the value vectors remain unchanged. The spatial coordinates are omitted for simplicity. This preserves relative positions within each segment but increases every cross-segment temporal distance by $\beta$, reducing ambiguity at the reference and video boundary.

\section{Experiments}

\subsection{Experiment Settings}

\textbf{Dataset.} We construct a benchmark from OpenVid-HD~\cite{nan2024openvid}. Unlike prior work~\cite{fei2025skyreels} that samples the reference directly from the target video (inviting shortcut learning via pixel copying), we edit the first frame of each ground-truth video while preserving subject identity, yielding harder reference and video pairs. This produces roughly 18{,}600 pairs; we randomly sample 5\% (933 videos) as our benchmark and use a VLM to generate an additional prompt per video for cross-video evaluation. The full pipeline is described in Sec.~\emph{Dataset Pipeline} of the supplementary material.

\noindent\textbf{Implementation detail.} We update all DiT blocks of Wan 2.1 T2V 1.3B~\cite{wan2025wan} while freezing the VAE, CLIP, and text encoders. Training uses AdamW with $\beta_1=0.9$, $\beta_2=0.95$, learning rate $1\times10^{-4}$, and $512\times512$ videos. We set the reconstruction-loss weight to $\lambda=f_r/f_v$ and the Gap-RoPE offset to $\beta=4$. At inference, we generate at $480\times832$ resolution with guidance scale 6 and 50 denoising steps. Additional training, inference, and evaluation details are provided in the supplementary material.

\noindent\textbf{Baselines.} We compare with Hunyuan Custom~\cite{hu2025hunyuancustomm}, Phantom~\cite{liu2025phantom}, VACE~\cite{jiang2025vace}, and ConsisID~\cite{yuan2025identity}. To match our scale, we use the 1.3B-parameter versions of Phantom and VACE. Since these baselines are not trained on augmented prompts, we evaluate them with the original OpenVid-HD prompt; all other hyperparameters use defaults.

\begin{table*}
\centering
\caption{Comparison across video quality, video-reference consistency, and inter-video consistency. OF Mag. uses RAFT~\cite{teed2020raft}, FVMD~\cite{liu2024fvmd} measures motion-distribution distance, and Dyn. Deg. is from VBench~\cite{huang2024vbench}. ``Base T2V'' is the unconditioned Wan 2.1 T2V 1.3B backbone and serves as a motion reference; reference-related metrics are not applicable. Bold marks the best reference-conditioned result.}
\vspace{-0.1in}
\resizebox{0.98\textwidth}{!}{
\begin{tabular}{l@{\hspace{10pt}}lccccccccccc}
\toprule
\multirow{2}{*}{\textbf{Method}} &
\multirow{2}{*}{\textbf{Params}} &
\multicolumn{5}{c}{\textit{Video Quality}} &
\multicolumn{3}{c}{\textit{Video-Reference Consistency}} &
\multicolumn{3}{c}{\textit{Inter-Video Consistency}} \\
\cmidrule(lr){3-7} \cmidrule(lr){8-10} \cmidrule(lr){11-13}
& & CLIP-I & Temp. Con. & OF Mag. & FVMD$\downarrow$ & Dyn. Deg.
& ArcFace & DINO-I & VLM-App.
& ArcFace & DINO-I & VLM-App. \\
\midrule
ConsisID        & 5B   & 0.2743 & 0.9258 & 4.21 & 612 & 0.78 & 0.5981 & 0.3188 & 0.5182 & 0.5682 & 0.2736 & 0.4143 \\
VACE            & 1.3B & 0.3012 & \textbf{0.9903} & 2.13 & 845 & 0.42 & 0.5489 & 0.4468 & 0.8219 & 0.5394 & 0.4103 & 0.8123 \\
Phantom         & 1.3B & 0.3095 & 0.9802 & \textbf{5.02} & 478 & \textbf{0.87} & 0.5636 & 0.4719 & 0.8821 & 0.5412 & 0.4419 & 0.8490 \\
Hunyuan Custom  & 13B  & 0.2897 & 0.9728 & 3.86 & 558 & 0.74 & 0.5571 & 0.4819 & 0.9360 & 0.5529 & 0.4782 & 0.9179 \\
\midrule
\textit{Base T2V} & \textit{1.3B} & \textit{0.3056} & \textit{0.9655} & \textit{5.41} & \textit{412} & \textit{0.89} & \textit{N/A} & \textit{N/A} & \textit{N/A} & \textit{N/A} & \textit{N/A} & \textit{N/A} \\
\midrule
\textbf{Ours}   & 1.3B & \textbf{0.3107} & 0.9881 & 4.89 & \textbf{432} & 0.86 & \textbf{0.6003} & \textbf{0.4824} & \textbf{0.9608} & \textbf{0.5943} & \textbf{0.4790} & \textbf{0.9457} \\
\bottomrule
\end{tabular}
}
\vspace{-0.1in}
\label{tab:eval_metrics}
\end{table*}

\subsection{Quantitative Evaluation}

We evaluate the proposed method using three categories of metrics that jointly assess video quality, video-reference consistency, and inter-video consistency.

\begin{figure*}[t]
    \centering
    \includegraphics[width=\linewidth]{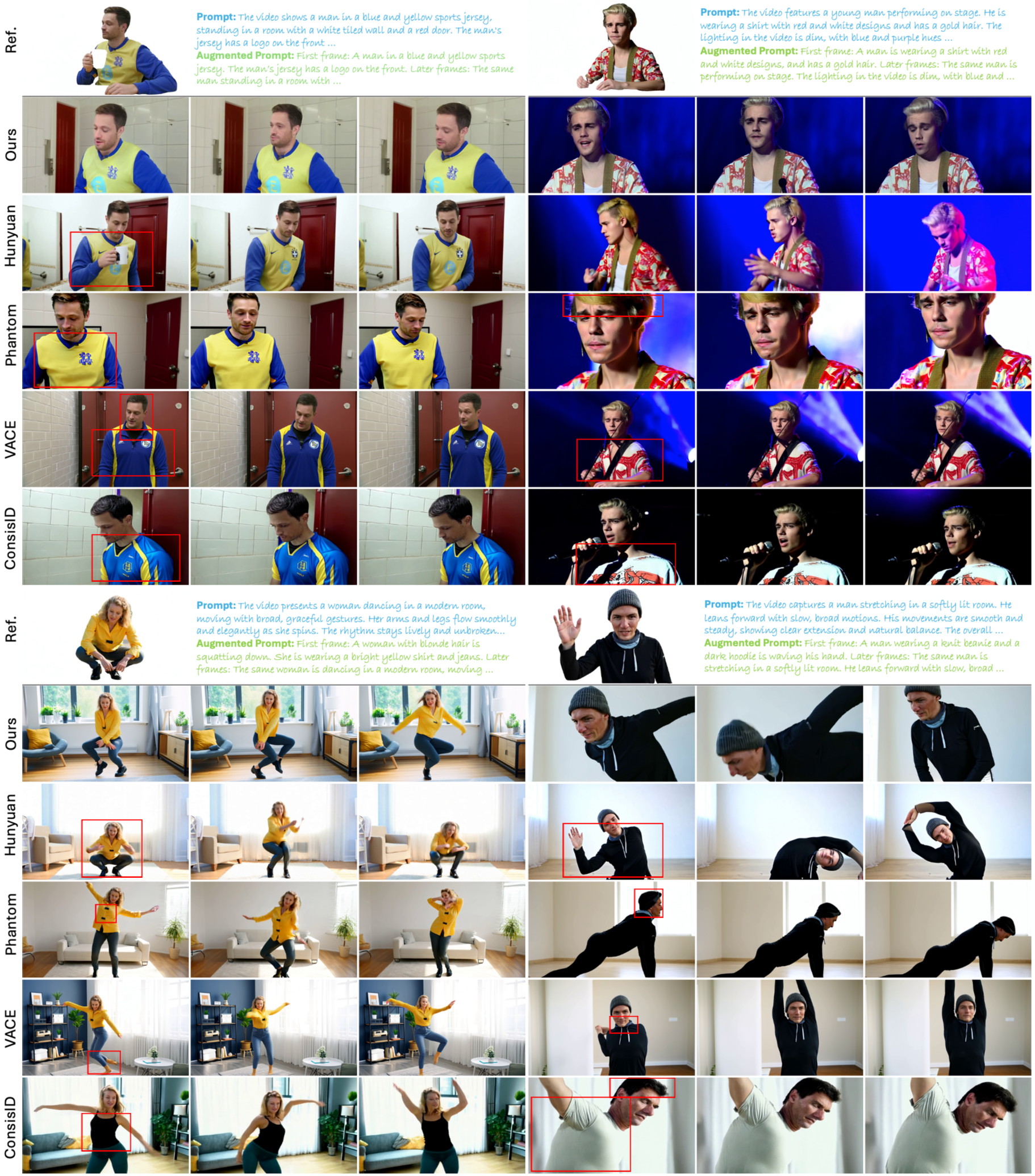}
    \caption{Qualitative comparison on fine-grained details, uncommon clothing, full-body generation, and large motion. Rows show the reference, ContextAnyone, Hunyuan Custom, Phantom, VACE, and ConsisID. The scenarios jointly test whether local patterns and holistic appearance remain stable across pose and motion changes. Red boxes mark appearance inconsistencies, which ContextAnyone reduces across the examples.}
    \label{fig:comparison}
\end{figure*}

\noindent\textbf{Video quality.} We report CLIP-I~\cite{hessel2021clipscore} for text and frame alignment, and temporal consistency as the cosine similarity between consecutive frames. Our method achieves the highest CLIP-I score. ConsisID and Hunyuan Custom either fail to preserve holistic appearance or overly copy the reference image (Fig.~\ref{fig:comparison}), yielding relatively low CLIP-I. Although VACE attains the highest temporal consistency, this stems from its tendency to generate nearly static videos, which artificially inflates frame-to-frame similarity; we verify this with the motion metrics discussed next.

A natural concern with reference-conditioned video generation is whether stronger appearance fidelity comes at the cost of motion diversity. We therefore report three motion-specific metrics in Table~\ref{tab:eval_metrics} and include the unconditioned Wan 2.1 1.3B backbone as a reference point. ContextAnyone remains close to this reference (OF Mag. 4.89 vs. 5.41, FVMD 432 vs. 412, Dyn. Deg. 0.86 vs. 0.89). Phantom produces slightly higher raw motion (OF Mag. 5.02, Dyn. Deg. 0.87) but lower identity and appearance consistency, whereas VACE has lower motion statistics (OF Mag. 2.13, Dyn. Deg. 0.42), which helps explain its high frame-to-frame similarity. Thus, ContextAnyone improves appearance preservation while retaining much of the base generator's motion range.

\noindent\textbf{Video-reference consistency.} We assess similarity between generated frames and the reference image using ArcFace~\cite{deng2019arcface} and DINO-I~\cite{caron2021emerging} features. To capture fine-grained person-specific cues such as clothing, hairstyle, and accessories that generic visual features miss, we further introduce \textbf{VLM-Appearance}: a VLM-based score in $[0,1]$ between the reference image and uniformly sampled generated frames; the full prompt template, aggregation, and design rationale are detailed in Sec.~\emph{VLM-Appearance Evaluation} of the supplementary material. Our method obtains the highest scores across all three metrics. A blind 30-participant study in the supplementary material prefers ContextAnyone to Phantom and VACE in 72.6\% and 82.3\% of comparisons, respectively. VLM-Appearance correlates with these judgments at Spearman $\rho=0.72$, compared with 0.38 for ArcFace and 0.51 for DINO-I.

\noindent\textbf{Inter-video consistency.} To assess consistency across different generated videos, we compare two videos generated from the same reference image under different prompts. We compute ArcFace and DINO-I similarity over frame pairs sampled from the two videos to measure identity and visual consistency across scenes, and additionally use VLM-Appearance to evaluate perceived character-level appearance consistency across videos. Our method outperforms all baselines across these metrics, demonstrating stronger robustness in maintaining consistent identity and appearance under varied prompts and scene changes.

\begin{figure}[t]
    \centering
    \includegraphics[width=\linewidth]{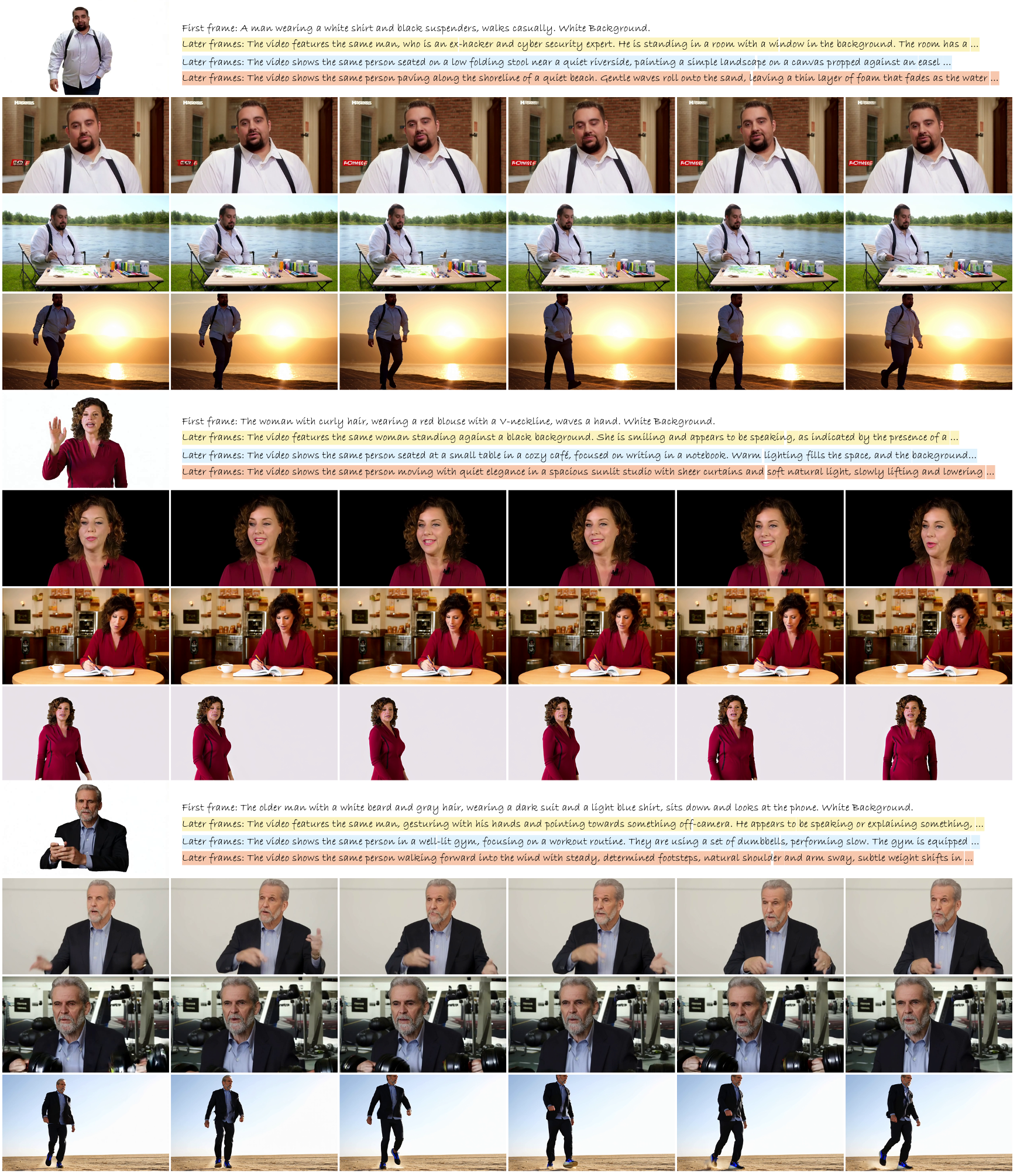}\\
    \vspace{-0.1in}
    \caption{Cross-video results for three subjects generated from the same reference under different prompts. ContextAnyone retains identity, hairstyle, and clothing details across scenes and motions.}
    \vspace{-0.1in}
    \label{fig:comparison2}
\end{figure}

\begin{figure}[t]
    \centering
    \includegraphics[width=\linewidth]{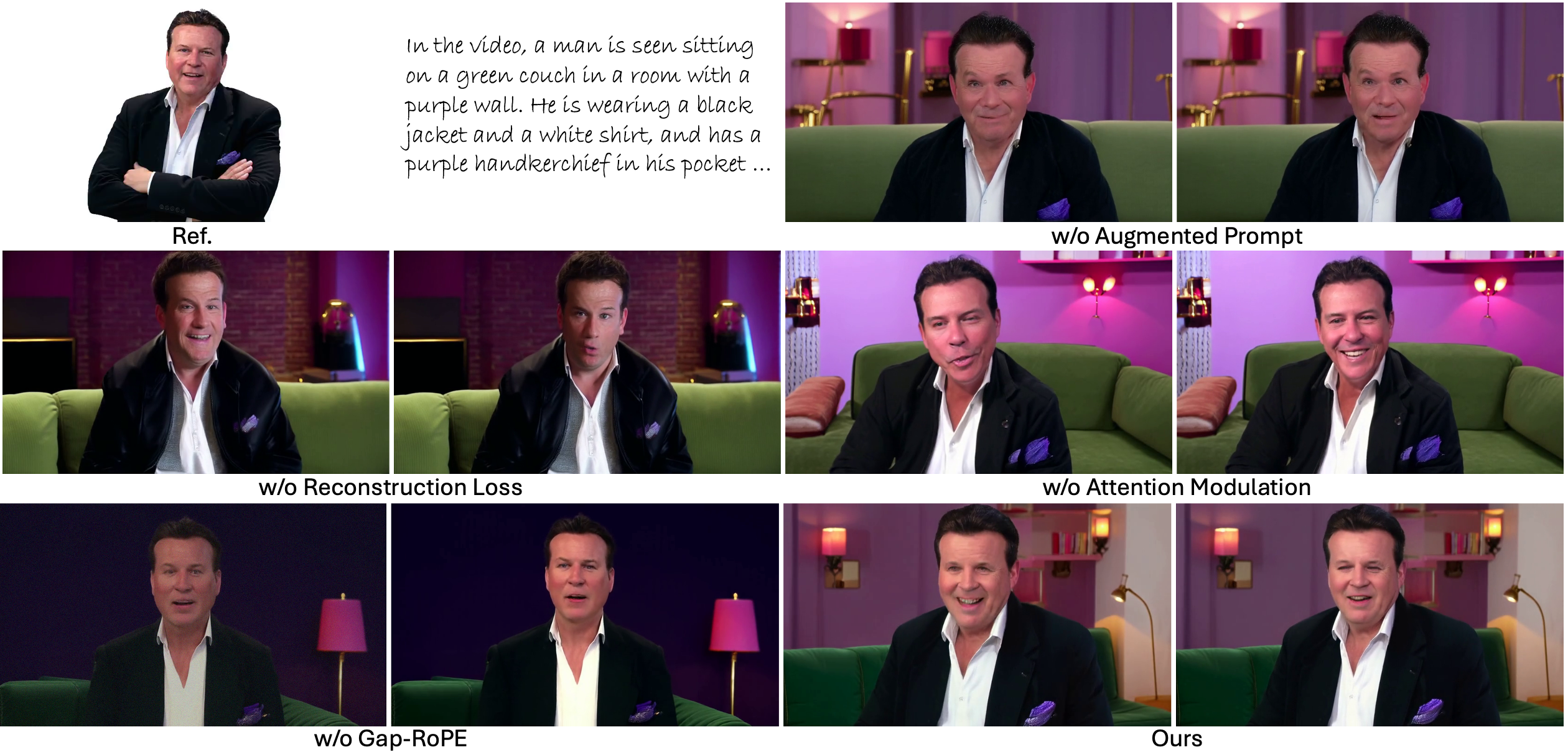}\\
    \vspace{-0.1in}
    \caption{Visualization of the ablation study.}
    \vspace{-0.15in}
    \label{fig:ablation}
\end{figure}

\subsection{Qualitative Evaluation}

Figure~\ref{fig:comparison} provides a qualitative comparison between our method and the baselines. For each group, we show the reference image (Ref.) followed by the results from different methods. Regions with inconsistencies or artifacts are highlighted with red boxes. Our method consistently preserves fine-grained facial features, including contours, skin tone, and expressions, while maintaining hairstyle structure and clothing textures across scenes. It demonstrates strong robustness to changes in background, lighting, and pose, achieving high visual coherence and accurate appearance transfer.

In the examples shown, Hunyuan Custom often remains close to the reference composition, while Phantom exhibits outfit discontinuities under pose and lighting changes. VACE shows larger deviations in facial regions, and ConsisID preserves facial identity more reliably than clothing and holistic appearance. Figure~\ref{fig:comparison2} provides additional cross-video results under different prompts.

\subsection{Ablation}
\label{sec:ablation}

\begin{table}
\centering
\caption{Ablation of the three architectural designs and prompt augmentation. The attention variant jointly removes Emphasize Attention and the directional self-attention mask. Removing each design reduces at least one evaluated metric.}
\vspace{-0.1in}
\resizebox{\linewidth}{!}{
\begin{tabular}{lccc}
\toprule
\textbf{Variant} & \textbf{Temp. Con} & \textbf{ArcFace} & \textbf{DINO-I} \\
\midrule
w/o Aug. Prompt & 0.9798 {\scriptsize\textcolor{red}{-0.0083}} & 0.5681 {\scriptsize\textcolor{red}{-0.0322}} & 0.4633 {\scriptsize\textcolor{red}{-0.0191}} \\
w/o Recon. Loss & 0.9721 {\scriptsize\textcolor{red}{-0.0160}} & 0.5271 {\scriptsize\textcolor{red}{-0.0732}} & 0.4203 {\scriptsize\textcolor{red}{-0.0621}} \\
w/o Attn. Design & 0.9756 {\scriptsize\textcolor{red}{-0.0125}} & 0.5592 {\scriptsize\textcolor{red}{-0.0411}} & 0.4189 {\scriptsize\textcolor{red}{-0.0635}} \\
w/o Gap-RoPE & 0.9410 {\scriptsize\textcolor{red}{-0.0471}} & 0.5814 {\scriptsize\textcolor{red}{-0.0189}} & 0.4622 {\scriptsize\textcolor{red}{-0.0202}} \\
\textbf{Ours} & \textbf{0.9881} & \textbf{0.6003} & \textbf{0.4824} \\
\bottomrule
\end{tabular}
}
\vspace{-0.15in}
\label{tab:ablation}
\end{table}

We ablate the design associated with each challenge in Sec.~\ref{sec:design_principles}, as well as the augmented prompt input. The attention variant jointly removes Emphasize Attention and the directional self-attention mask; the experiment therefore evaluates the combined attention design rather than isolating its two parts. Figure~\ref{fig:ablation} and Table~\ref{tab:ablation} show that removing any of the three architectural designs reduces at least one evaluated metric.

\textit{Joint reconstruction.} Removing $\mathcal{L}_\text{ref}$ produces the largest ArcFace decrease among the evaluated variants ($-0.0732$) and visibly distorts facial features. DINO-I also decreases by $0.0621$, while Temp.~Con. changes less ($-0.0160$), showing that explicit supervision on the reference branch benefits both facial identity and broader appearance.

\textit{Asymmetric attention.} Jointly removing Emphasize Attention and the directional mask produces the largest DINO-I decrease among the variants ($-0.0635$); the purple pocket square also becomes less consistent across frames. ArcFace and Temp.~Con. decrease by $0.0411$ and $0.0125$, respectively. These results support the combined attention design, but do not isolate the contribution of its two constituent modifications.

\textit{Gap-RoPE.} Removing Gap-RoPE produces the largest Temp.~Con. decrease among the variants ($-0.0471$) and introduces visible noise artifacts in early frames, while ArcFace and DINO-I decrease by $0.0189$ and $0.0202$, respectively. This indicates that positional separation is particularly useful for temporal consistency in our setting.

Because the three metrics have different scales, we do not compare raw decreases across columns. Within each metric, however, the largest decrease is associated with a different removal: joint reconstruction for ArcFace, the combined attention design for DINO-I, and Gap-RoPE for Temp.~Con. We interpret this approximately diagonal empirical pattern as evidence of complementary contributions, rather than proof of statistical independence or necessity. Removing prompt augmentation also reduces all three metrics, supporting its role as an input-side complement.

\noindent\textbf{Augmented prompt fairness.} We re-evaluate all baselines with the augmented prompt format used by our method. The format does not improve the evaluated baselines (e.g., Hunyuan Custom decreases by 0.010 on VLM-Appearance and ConsisID by 0.027 on ArcFace), while ContextAnyone also improves over its original-prompt variant in Table~\ref{tab:ablation}. Thus, prompt augmentation provides an additional gain for ContextAnyone but does not explain its improvement under the original-prompt comparison. Full per-baseline results are provided in the supplementary material.

\section{Conclusion}

We studied holistic appearance preservation for single-reference, single-character video generation with token-concatenation DiTs. ContextAnyone treats the reference as an explicit appearance anchor through joint reconstruction, asymmetric attention, and Gap-RoPE. Experiments on Wan 2.1 1.3B demonstrate improved holistic appearance consistency over the evaluated public baselines while retaining the motion characteristics of the base generator. Our evaluation is limited to a single backbone and single-person references, and distortions can still occur under large motions or for unobserved body regions. Future work will explore other backbones as well as multi-reference and multi-character settings.


{
    \small
    \bibliographystyle{ieeenat_fullname}
    \bibliography{main}
}

\end{document}